\documentclass[conference]{IEEEtran}
\IEEEoverridecommandlockouts
\usepackage{array} 
\usepackage{booktabs}
\usepackage{bm}
\usepackage{cite}
\usepackage{soul}
\usepackage{xcolor}
\usepackage{siunitx}
\usepackage{balance}
\usepackage{subfig}
\usepackage[hidelinks]{hyperref}
\usepackage{colortbl}
\usepackage{graphicx}
\usepackage{makecell}
\usepackage{multirow}
\usepackage{float}
\usepackage{textcomp}
\usepackage{mathtools}
\usepackage{algorithmic}
\usepackage[T1]{fontenc}
\usepackage{threeparttable}
\usepackage{amsmath,amssymb,amsfonts}
\usepackage[color=blue!40]{todonotes}
\usepackage[ruled,algoruled]{algorithm2e}
\hypersetup{
  bookmarksnumbered,
  colorlinks=false,       
  pdfborder={0 0 1}       
}%

\soulregister\footnote7
\soulregister\eqref7
\soulregister\cite7
\soulregister\ref7
\graphicspath{{./}{../figure/}}

\begin{document}

\title{
  EKF-Based Fusion of Wi-Fi/LiDAR/IMU for Indoor Localization and Navigation}%
\author{%
  \IEEEauthorblockN{%
    Zeyi Li$^{1,2}$, Zhe Tang$^{3}$, Kyeong Soo Kim$^{1}$, Sihao Li$^{4}$, and
    Jeremy S. Smith$^{2}$}%
  \IEEEauthorblockA{%
    $^{1}$School of Advanced Technology, Xi'an Jiaotong-Liverpool University, Suzhou, 215123, P.R. China.\\
    $^{2}$Department of Electrical Engineering and Electronics, University of Liverpool, Liverpool, L69 3GJ, U.K.\\
    $^{3}$Zhejiang University of Technology, Hangzhou, 310014, P.R. China.\\
    $^{4}$School of Artificial Intelligence, Suzhou Vocational Institute of Industrial Technology, Suzhou, 215104, P.R. China.\\
    Email: Zeyi.Li21@alumni.xjtlu.edu.cn, timtangprc@gmail.com, Kyeongsoo.Kim@xjtlu.edu.cn, 01177@siit.edu.cn,\\
    J.S.Smith@liverpool.ac.uk}%
  \thanks{%
    Z. Tang and S. Li were with the School of Advanced Technology at Xi'an
    Jiaotong-Liverpool University and the Department of Electrical Engineering
    and Electronics at the University of Liverpool when this work was done.

    \vspace{0.1cm}
    \noindent%
    * Corresponding author: Kyeongsoo.Kim@xjtlu.edu.cn}%
}%

\maketitle

\begin{abstract}
  Conventional Wi-Fi received signal strength indicator (RSSI) fingerprinting
  cannot meet the growing demand for accurate indoor localization and navigation
  due to its lower accuracy, while solutions based on light detection and
  ranging (LiDAR) can provide better localization performance but is limited by
  their higher deployment cost and complexity. To address these issues, we
  propose a novel indoor localization and navigation framework integrating Wi-Fi
  RSSI fingerprinting, LiDAR-based simultaneous localization and mapping (SLAM),
  and inertial measurement unit (IMU) navigation based on an extended Kalman
  filter (EKF). Specifically, coarse localization by deep neural network
  (DNN)-based Wi-Fi RSSI fingerprinting is refined by IMU-based dynamic
  positioning using a Gmapping-based SLAM to generate an occupancy grid map and
  output high-frequency attitude estimates, which is followed by EKF
  prediction-update integrating sensor information while effectively suppressing
  Wi-Fi-induced noise and IMU drift errors. Multi-group real-world experiments
  conducted on the IR building at Xi'an Jiaotong-Liverpool University
  demonstrates that the proposed multi-sensor fusion framework suppresses the
  instability caused by individual approaches and thereby provides stable
  accuracy across all path configurations with mean two-dimensional (2D) errors
  ranging from \SI{0.2449}{\m} to \SI{0.3781}{\m}. In contrast, the mean 2D
  errors of Wi-Fi RSSI fingerprinting reach up to \SI{1.3404}{\m} in areas with
  severe signal interference, and those of LiDAR/IMU localization are between
  \SI{0.6233}{\m} and \SI{2.8803}{\m} due to cumulative drift.
\end{abstract}


\section{Introduction}
\label{sec:intro}
With the widespread deployment of indoor service robots, intelligent logistics
systems, and mobile terminals in smart buildings, high-accuracy indoor
localization technology has become a key enabler for space perception and
autonomous navigation. However, traditional global positioning systems (GPS) are
severely obstructed in indoor environments, where signal attenuation and
multipath interference render them virtually
useless~\cite{wahab2022indoor}. Therefore, researchers have turned to indoor
localization methods based on wireless signals and sensor fusion to accurately
track moving objects in complex spatial environments.

Currently, mainstream indoor localization methods include Wi-Fi fingerprinting,
ultra wide band (UWB) ranging, Bluetooth near-field localization, and inertial
measurement unit (IMU) navigation~\cite{guo2019survey}. Wi-Fi fingerprinting has
been favored in many application scenarios because it does not require the
installation of any new infrastructure or the modification of existing devices,
which results in lower deployment cost. However, this method relies heavily on a
pre-collected received signal strength indicator (RSSI) database, which is
susceptible to environmental changes and device differences, resulting in poor
stability and versatility~\cite{gustafsson2003positioning,yiu2017wireless}. IMU
can quickly estimate the orientation (attitude) and motion of a body and thereby
is widely used in fast localization tasks in dynamic environments. Classic IMU
localization methods include the double integration method based on physical
models, the pedestrian dead reckoning (PDR) method based on path estimation, the
Kalman filter algorithm, and fusion methods with other
sensors~\cite{yan2018ridi,jirawimut2003method,caron2006gps,dehzangi2017imu}. However,
the localization performance of these methods degrades quickly as the errors in
the acceleration and angular velocity integration processes, especially the
drift effect, accumulate~\cite{wu2015indoor}. Light detection and ranging
(LiDAR) offers extremely high spatial ranging accuracy and is commonly used for
map construction and pose estimation in simultaneous localization and mapping
(SLAM) systems \cite{he2020integrated}. However, its high cost and environmental
dependencies limit its scalability for large-scale deployment.

Due to the limitations of the single-sensor solutions mentioned above,
multi-sensor fusion methods have garnered widespread attention, in recent years,
because joint modeling and filter estimation of different sensor information can
greatly improve the robustness and localization accuracy of a
system~\cite{jo2011interacting}. Among them, the extended Kalman filter (EKF),
an established method for processing non-linear state estimation, has been
widely used in robot perception and navigation systems due to its advantages of
strong real-time performance, clear structure, and simple engineering
implementation~\cite{ribeiro2004kalman,einicke2002robust}. However, current
research is mostly limited to two-source fusion schemes such as Wi-Fi/PDR or
UWB/IMU~\cite{liu2021kalman}.

We propose a multi-sensor fusion framework for accurate and dynamic indoor
localization and navigation under complex indoor environments, where Wi-Fi RSSI
fingerprinting and LiDAR SLAM enhanced by IMU, are integrated based on EKF. We
carry out a comparative analysis of the performance of the proposed framework
through comprehensive experiments in the IR building of Xi'an Jiaotong-Liverpool
University (XJTLU), whose results demonstrate that the proposed framework
consistently outperforms Wi-Fi and LiDAR/IMU-based solutions, achieving superior
error control and trajectory smoothness performance.

The rest of the paper is organized as follows: Section~\ref{sec:related_work}
reviews related work. Section~\ref{sec:methodology} introduces the proposed
EKF-based indoor localization framework. Section~\ref{sec:exp-results} presents
experimental results and compares the performance of different localization
methods across multiple environments. Section~\ref{sec:conclusions} presents the
conclusions from our work.

\section{Related Work}
\label{sec:related_work}
\subsection{EKF in Indoor Localization}
\label{sec:ekf-indoorloc}
EKF is a recursive state estimation algorithm widely used in indoor localization
systems~\cite{feng2020kalman}. It can fuse asynchronous, noisy, and multiple
sensor observations in environments with non-linear dynamic characteristics and
estimate user location information in real time within a probabilistic
framework. Compared to traditional linear Kalman filters, the EKF linearizes
non-linear systems using a first-order Taylor expansion, enabling it to handle
non-linear motion and observation models. This feature makes it more adaptable
and robust in indoor environments with complex user trajectories and uncertain
signal propagation paths~\cite{barrau2015non}.

The performance of indoor localization systems based on Wi-Fi RSSIs and channel
state information (CSI) can be significantly affected due to multipath effects,
channel attenuation, and dynamic changes in the
environment~\cite{laoudias2018survey}. EKF can filter and correct noisy
observations by modeling state transitions and observation processes. To further
improve system performance, researchers have recently combined EKF with inertial
and environmental perception modalities such as IMU and LiDAR to fuse absolute
localization and relative motion information, thereby enhancing the continuity
and stability of trajectory estimation~\cite{laoudias2018survey}.

In addition, some studies have introduced deep neural networks (DNNs) for
preprocessing Wi-Fi fingerprint data (e.g., using autoencoders for feature
denoising and reconstruction to reduce the impact of environmental factors on
signal distribution). Subsequently, Wi-Fi features enhanced by DNNs are jointly
input into the EKF framework along with IMU data, enabling the system to
simultaneously perceive static scene features and dynamic motion information
during state updates, thereby effectively mitigating initial heading errors and
short-term drift issues~\cite{zhou2024wio}.

\subsection{Multi-Sensor Fusion in Indoor Localization}
\label{sec:multi-sensor-fusion-indoorloc}
Conventional indoor localization methods based on a single signal source are
susceptible to dynamic changes and channel interference in complex scenarios,
reducing localization accuracy and system stability. For this reason,
multi-sensor fusion has received widespread attention in recent years as a
promising solution to improve the accuracy and robustness of indoor localization
systems. This technology integrates heterogeneous sensor data from different
modalities to achieve more accurate, continuous, and robust estimates of target
locations, effectively addressing uncertainties in dynamic
environments~\cite{tang2023comparative}.

From the perspective of fusion architecture, multi-sensor fusion methods can be
categorized into data-level, feature-level, and decision-level
fusion~\cite{zafari2019survey}. Data-level fusion directly integrates raw
observation data from sensors, exploiting the most complete
information. However, it has high requirements for time synchronization and
noise robustness. Feature-level fusion, on the other hand, integrates the
extracted features of each sensor's data, balancing computational efficiency and
localization accuracy. Decision-level fusion weights or filters the results of
the inference by a subsystem dedicated to each sensor, which is suitable for
distributed architectures and heterogeneous device environments. Examples of
recent development in multi-sensor fusion in indoor localization include the
complementary fusion of Wi-Fi and IMU to improve short-term stability in
scenarios without GPS~\cite{liu2019fusion}, cooperative localization of UWB and
visual sensors (such as cameras or LiDAR) to achieve centimeter-level accuracy
in industrial and robotics scenarios~\cite{gu2020fusion}, and multi-modal fusion
strategies based on end-to-end neural networks to improve robust perception
capabilities in complex environments through non-linear
modeling~\cite{zhou2022deep}.

Although multi-sensor fusion can improve localization accuracy and stability, it
still faces many challenges~\cite{zafari2019survey}. For example, issues such as
data synchronization between multi-source sensors, coordinate system alignment,
and scale consistency have not been fully resolved. Likewise, the limited
computational power of mobile devices imposes serious challenges to the design
of real-time, energy-efficient fusion algorithms. In addition, data
heterogeneity, outlier handling, and the complexity of system integration
further restrict the feasibility of such methods for large-scale deployment.

\section{Methodology}
\label{sec:methodology}
\begin{figure}[!htb]
  \centering%
  \includegraphics[width=.8\linewidth]{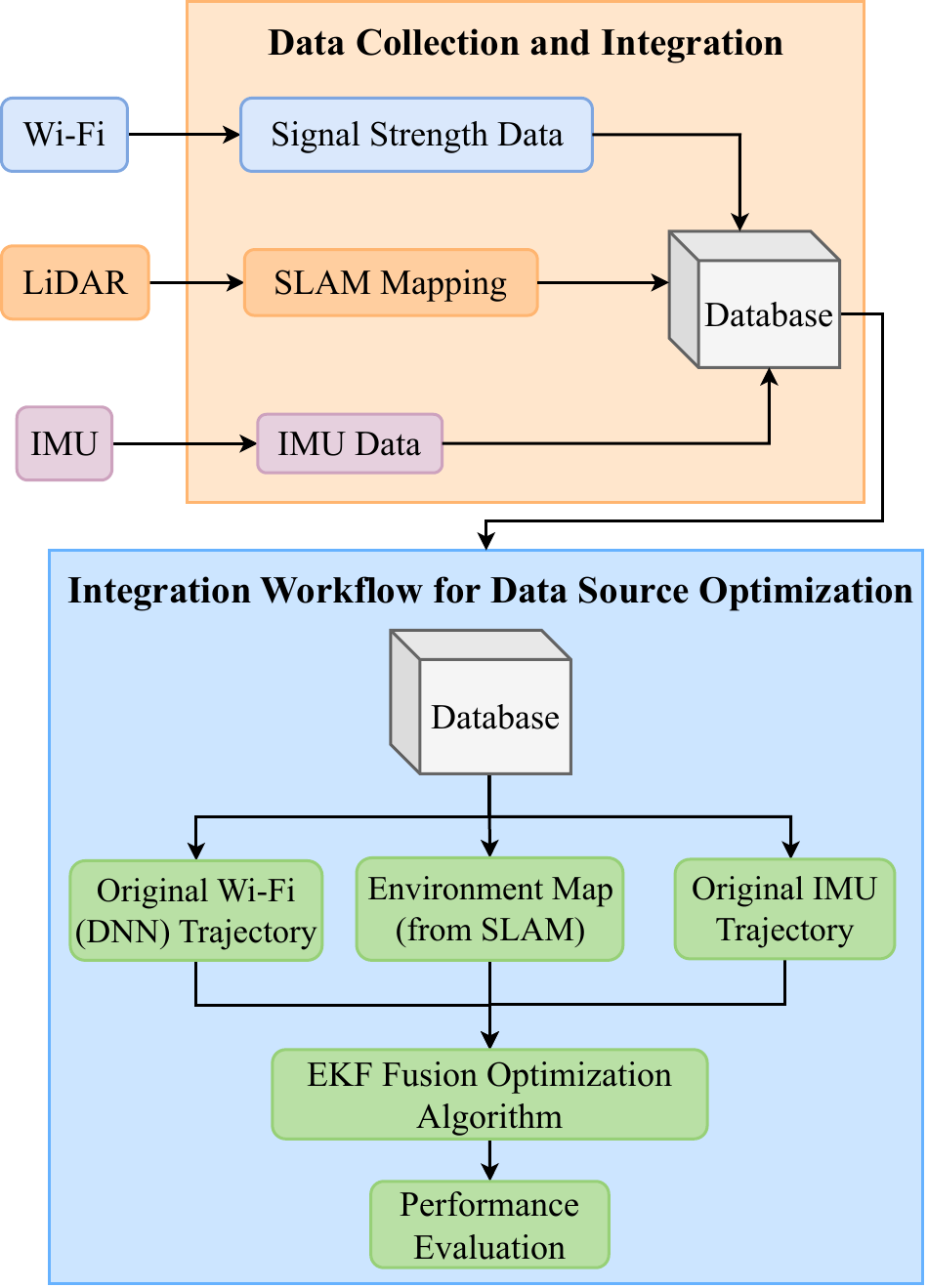}
  \caption{Overall architecture and information processing flow of the
    multi-sensor indoor localization and navigation system.}
  \label{fig:overview_plus}
\end{figure}

We construct a high-accuracy and robust indoor localization system that
integrates three types of sensor information: Wi-Fi RSSIs, LiDAR-based SLAM, and
IMU three-axis acceleration and angular velocity. It achieves state estimation
and real-time correction through EKF. The method covers four significant steps:
Data collection, modeling and training, state estimation, and performance
evaluation. The overview of the proposed system is shown in
\autoref{fig:overview_plus}.

\subsection{Data Processing and Integration Process}
\label{sec:data-proc-integr}
\begin{figure}[h]
  \centering
  \includegraphics[width=\linewidth]{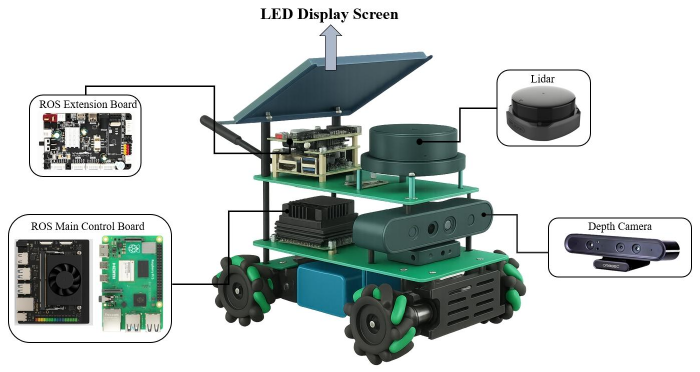}
  \caption{AGV platform for the proposed system.}
  \label{fig:agv}
\end{figure}

The proposed system is implemented on the autonomous guided vehicle (AGV)
platform shown in \autoref{fig:agv}. This platform integrates an NVIDIA Jetson
computing module, a Wi-Fi receiver module, a two-dimensional (2D) LiDAR, and an
IMU, enabling synchronous multi-source data collection. The raw data collected
is stored and preprocessed after being aligned with a unified timestamp,
including missing value imputation, outlier removal, and normalization
operations, to ensure the consistency and fusibility of data from different
modalities.

During operation, the Wi-Fi module collects RSSI vectors from surrounding access
points (APs) to construct a fingerprint database, the LiDAR provides SLAM
mapping and pose trajectory estimation, and the IMU makes high-frequency
measurement of linear acceleration and angular velocity data. All data is
uploaded to the edge computing unit for local processing and model prediction.

\subsection{DNN-Based Wi-Fi RSSI Fingerprinting}
\label{sec:wi-fi-fingerprinting}
To improve the localization accuracy of Wi-Fi RSSI fingerprinting in static
areas, we use a DNN regression model to estimate the 2D coordinates of the AGV
through non-linear mapping between a high-dimensional RSSI vector and physical
space 2D coordinates under a supervised learning framework.

\begin{figure}[h]
  \centering
  \includegraphics[width=.8\linewidth]{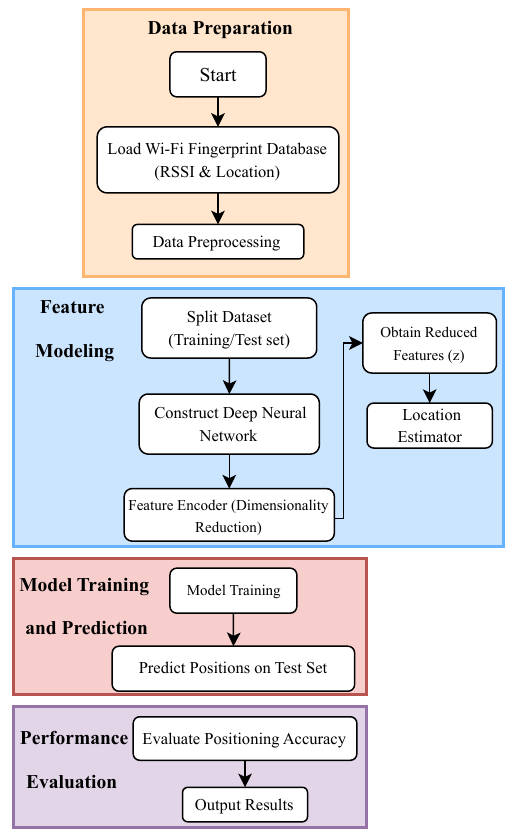}
  \caption{Workflow of the Wi-Fi RSSI fingerprinting based on a DNN model.}
  \label{fig:wifi_dnn_workflow}
\end{figure}

As shown in~\autoref{fig:wifi_dnn_workflow}, the overall process of the Wi-Fi
localization module is divided into four stages: data preparation, feature
modeling, model training and prediction, and performance evaluation. Initially,
the system loads a pre-built fingerprint database containing RSSI vectors
collected at different reference points (RPs) and their corresponding 2D
coordinates. To improve modeling quality and training stability, the original
data undergoes preprocessing operations such as missing value filling, feature
normalization, and outlier removal, and is then divided into training and test
datasets.

A multi-layer feedforward neural network is used for the feature modeling stage
to extract potential spatial distribution patterns from high-dimensional RSSI
vectors. \autoref{tab:wifi_dnn_structure} summarizes the parameter values of the
DNN model. The network consists of seven linear transformation units, each with
a rectified linear unit (ReLU) activation function and a dropout regularization
mechanism with a dropout rate of 0.2 to suppress overfitting and improve
generalization ability. The model input is 219-dimensional RSSI features
(corresponding to the number of valid APs), and the output is 2D localization
coordinates $(x,y)$.
\begin{table}[ht]
  \centering
  \caption{DNN parameter values for 2D-coordinate regression.}
  \label{tab:wifi_dnn_structure}
  \begin{tabular}{ccccc}
    \toprule
    \multirow{2}{*}{\textbf{Layer}} & \multirow{2}{*}{\textbf{Type}} & \textbf{Input} & \textbf{Output} & \textbf{Activation /} \\
                                    & & \textbf{Dimension} & \textbf{Dimension} & \textbf{Dropout} \\
    \midrule
    1 & Linear & 219  & 109 & ReLU / Dropout(0.2) \\
    2 & Linear & 109  & 73  & ReLU / Dropout(0.2) \\
    3 & Linear & 73   & 54  & ReLU / Dropout(0.2) \\
    4 & Linear & 54   & 109 & ReLU / Dropout(0.2) \\
    5 & Linear & 109  & 109 & ReLU / Dropout(0.2) \\
    6 & Linear & 109  & 109 & ReLU / Dropout(0.2) \\
    7 & Linear & 109  & 2   & None \\
    \bottomrule
  \end{tabular}
\end{table}

The DNN model is trained using mean squared error (MSE) as the loss function and
the Adam optimizer for parameter learning. During the testing phase, the trained
model predicts the 2D coordinates of given RSSI vectors, whose localization
performance is evaluated based on the Euclidean distances between predicted and
actual coordinates. The final results show that the DNN model has strong
robustness and generalization capabilities and can be used as global
localization prior information in the fusion phase to effectively support
continuous trajectory estimation in subsequent multi-source fusion.

\subsection{LiDAR/IMU-Based Dynamic Positioning}
\label{sec:imu-assisted-dynamic}
To improve the localization continuity and robustness of the system in complex
dynamic environments, we present a LiDAR-based SLAM method enhanced by IMU
data. Although the DNN-based Wi-Fi RSSI fingerprinting can provide high accuracy
in static environments, its generalization ability decreases in scenarios with
severe signal fluctuations, multipath propagation, serious obstructions, and
high fingerprint database update costs, making it difficult to achieve stable
and continuous location estimation.

To address the issues of the DNN-based Wi-Fi RSSI fingerprinting, we propose a
SLAM method based on multi-sensor observations without prior maps, enabling
real-time environment mapping and autonomous localization. The proposed method
is widely used in mobile robots and indoor navigation systems, with the key idea
of estimating the mobile platform's trajectory while dynamically updating the
spatial representation of the environment, thereby achieving the joint inference
of spatial and motion states.

The proposed method utilizes a Gmapping algorithm based on the Rao-Blackwellized
particle filtering to integrate 2D laser radar scan data with wheel odometer
information, thereby generating a closed-loop, consistent, high-resolution 2D
environment map through a probabilistic grid mapping model, while outputting a
high-precision pose trajectory estimate~\cite{grisetti2007improved}. Compared
with other SLAM methods, Gmapping has advantages such as, mature implementation,
low computational resource overhead, and strong closed-loop detection
capabilities, making it particularly suitable for real-time mapping and
localization tasks in small and medium scale indoor scenes.

To further improve the system's dynamic response capability under challenging
conditions such as rapid motion, sharp turns, and partial sensor occlusion, we
introduce an IMU as an auxiliary sensor. A typical IMU equipped with three-axis
accelerometer and gyroscope can measure three-dimensional (3D) linear
acceleration $\bm{a}{=}[a_{x}~a_{y}~a_{z}]^{\intercal}$ and angular velocity
$\bm{\omega}{=}[\omega_{x}~\omega_{y}~\omega_{z}]^{\intercal}$, which can be
modeled as follows:
\begin{equation}
  \label{eq:imu-measurement}
  \begin{split}
    \bm{a} & = \bm{a}_{\text{true}} + \bm{b}_a + \bm{n}_a, \\
    \bm{\omega} & = \bm{\omega}_{\text{true}} + \bm{b}_\omega + \bm{n}_\omega,
  \end{split}
\end{equation}
where $\bm{a}_{\text{true}}$ and $\bm{\omega}_{\text{true}}$ denote the true
values of 3D acceleration and angular velocity, $\bm{b}_{a}$ and
$\bm{b}_{\omega}$ the measurement biases, and $\bm{n}_{a}$ and $\bm{n}_{\omega}$
the measurement noise, respectively. The measurement biases and noise
in~\eqref{eq:imu-measurement} are modeled and corrected during the fusion stage
using the EKF to improve localization stability and accuracy.

Note that the high sampling frequency of the IMU can also effectively compensate
for the shortcomings of LiDAR in short-term motion estimation, improving the
localization continuity and robustness of the system in highly dynamic
environments.

\subsection{EKF-Based Multi-Sensor Fusion Based on Wi-Fi, LiDAR, and IMU}
\label{sec:multi-sensor-fusion}
EKF is a basis of the proposed sensor fusion framework for indoor localization,
which integrates the low-frequency global observations from Wi-Fi RSSI
fingerprinting and the high-frequency motion data from an IMU. The objective is
to produce accurate, continuous, and drift-resilient estimates of the platform's
state in real time.

\subsubsection{State Transition and Observation Models}
\label{sec:ekf-models}
The state vector and the control input of a system at time step $k$ are defined
as follows:
\begin{equation}
  \label{eq:ekf-models}
  \begin{split}
    \bm{X}_k & = \begin{bmatrix} x_k & y_k & \theta_k & v_k & \omega_k \end{bmatrix}^{\intercal}, \\
    \bm{u}_k & = \begin{bmatrix} a_k & \omega_k \end{bmatrix}^{\intercal},
  \end{split}
\end{equation}
where $x_k$ and $y_k$ represent the $x$ and $y$ coordinates of a 2D position,
$\theta_k$ is the heading angle, $v_k$ is the linear velocity, $\omega_k$ is the
angular velocity, $a_k$ is the measured linear acceleration, and $\omega_k$ is
the measured angular velocity, respectively. The state vector $\bm{X}_{k}$
captures both kinematic and orientation dynamics for modeling the mobile
platform behavior, while the control input $\bm{u}_{k}$ represents the
measurement data from the IMU.

\subsubsection{Prediction}
\label{sec:ekf-prediction}
Based on the prior state and control input, the predicted state
$\hat{\bm{X}}_{k|k-1}$ can be obtained assuming constant acceleration:
\begin{equation}
  \hat{\bm{X}}_{k|k-1} =
  \begin{bmatrix}
    x_{k-1} + v_{k-1} \Delta t \cos\theta_{k-1} \\
    y_{k-1} + v_{k-1} \Delta t \sin\theta_{k-1} \\
    \theta_{k-1} + \omega_{k-1} \Delta t \\
    v_{k-1} + a_k \Delta t \\
    \omega_k
  \end{bmatrix},
\end{equation}
where ${\Delta}t$ denotes the sampling interval.

The corresponding covariance matrix is propagated using a linearized transition
model:
\begin{equation}
  \bm{P}_{k|k-1} = \bm{F}_k \bm{P}_{k-1} \bm{F}_k^{\intercal} + \bm{Q}_k,
\end{equation}
where $\bm{F}_k$ is the Jacobian matrix of the transition function concerning
$\bm{X}_k$, and $\bm{Q}_k$ is the process noise covariance capturing system
model uncertainties.

\subsubsection{Update}
\label{sec:ekf-update}
The predicted state $\hat{\bm{X}}_{k|k-1}$ can be updated based on a
newly-observed 2D location from Wi-Fi RSSI fingerprinting, which is modeled by
\begin{equation}
  \bm{z}_k = \bm{H}_k \bm{X}_k + \bm{v}_k,
\end{equation}
and
\begin{equation}
  \bm{z}_k =
  \begin{bmatrix}
    x_k^{\text{obs}} \\
    y_k^{\text{obs}}
  \end{bmatrix},
  \quad
  \bm{H}_k =
  \begin{bmatrix}
    1 & 0 & 0 & 0 & 0 \\
    0 & 1 & 0 & 0 & 0
  \end{bmatrix},
  \quad
  \bm{v}_k \sim \mathcal{N}(\bm{0}, \bm{R}_k),
\end{equation}
where $\bm{z}_k$ represents the newly-observed 2D location, $\bm{H}_k$ denotes
the observation matrix that maps the full state vector $\bm{X}_k$ to the
measured position components, and $\bm{v}_k$ is a zero-mean Gaussian noise with
covariance matrix $\bm{R}_k$, respectively.

The Kalman gain, which balances the trust between prediction and observation, is
given by
\begin{equation}
  \bm{K}_k = \bm{P}_{k|k-1} \bm{H}_k^{\intercal} \left(\bm{H}_k \bm{P}_{k|k-1} \bm{H}_k^{\intercal} + \bm{R}_k\right)^{-1}.
\end{equation}
Using the innovation term, the state can be updated as follows:
\begin{equation}
  \hat{\bm{X}}_k = \hat{\bm{X}}_{k|k-1} + \bm{K}_k \left(\bm{z}_k - \bm{H}_k \hat{\bm{X}}_{k|k-1}\right).
\end{equation}
The covariance matrix, too, is updated as follows:
\begin{equation}
  \bm{P}_k = (\bm{I} - \bm{K}_k \bm{H}_k) \bm{P}_{k|k-1}.
\end{equation}

This EKF formulation supports the recursive fusion of asynchronous,
heterogeneous sensor data. By leveraging the stability of Wi-Fi RSSI
fingerprinting and the temporal resolution of the IMU signals, the proposed approach
produces continuous and accurate indoor localization estimates that are robust
to signal fluctuation, drift, and environmental noise.

\section{Experimental Results}
\label{sec:exp-results}
We conduct experiments in the IR building at XJTLU, where we estimate the
location of the AGV shown in \autoref{fig:agv} moving along the corridors of the
6th, 7th, and 8th floors. As the AVG moves along predefined trajectories, it
collects data using the equipped Wi-Fi, LiDAR, and IMU modules.

For comparative analyses, we investigate the localization performance of the
following three methods:
\begin{itemize}
\item DNN-based Wi-Fi RSSI fingerprinting (Wi-Fi).
\item LiDAR/IMU-based dynamic positioning (LiDAR/IMU).
\item EKF-based multi-sensor fusion based on Wi-Fi, LiDAR, and IMU (EKF).
\end{itemize}

As a performance metric, we use \textit{mean 2D error}, which is defined over
the test positions of a given trajectory as follows:
\begin{multline}
  \label{eq:2d-error}
  \text{Mean 2D Error} = \\
  \frac{1}{N} \sum_{i=1}^{N} \sqrt{\left(x^{\text{true}}_{i} -
      x^{\text{pred}}_{i}\right)^{2} + \left(y^{\text{true}}_{i} - y^{\text{pred}}_{i}\right)^{2}},
\end{multline}
where $N$ is the number of the test positions, and
$(x^{\text{true}}_i,y^{\text{true}}_i)$ and
$(x^{\text{pred}}_i,y^{\text{pred}}_i)$ are 2D coordinates of true and predicted
positions, respectively.

%


\subsection{Multi-Floor Trajectory Estimation}
\label{sec:multi-floor-exp}
\begin{figure}[h]
  \centering
  \includegraphics[width=\linewidth]{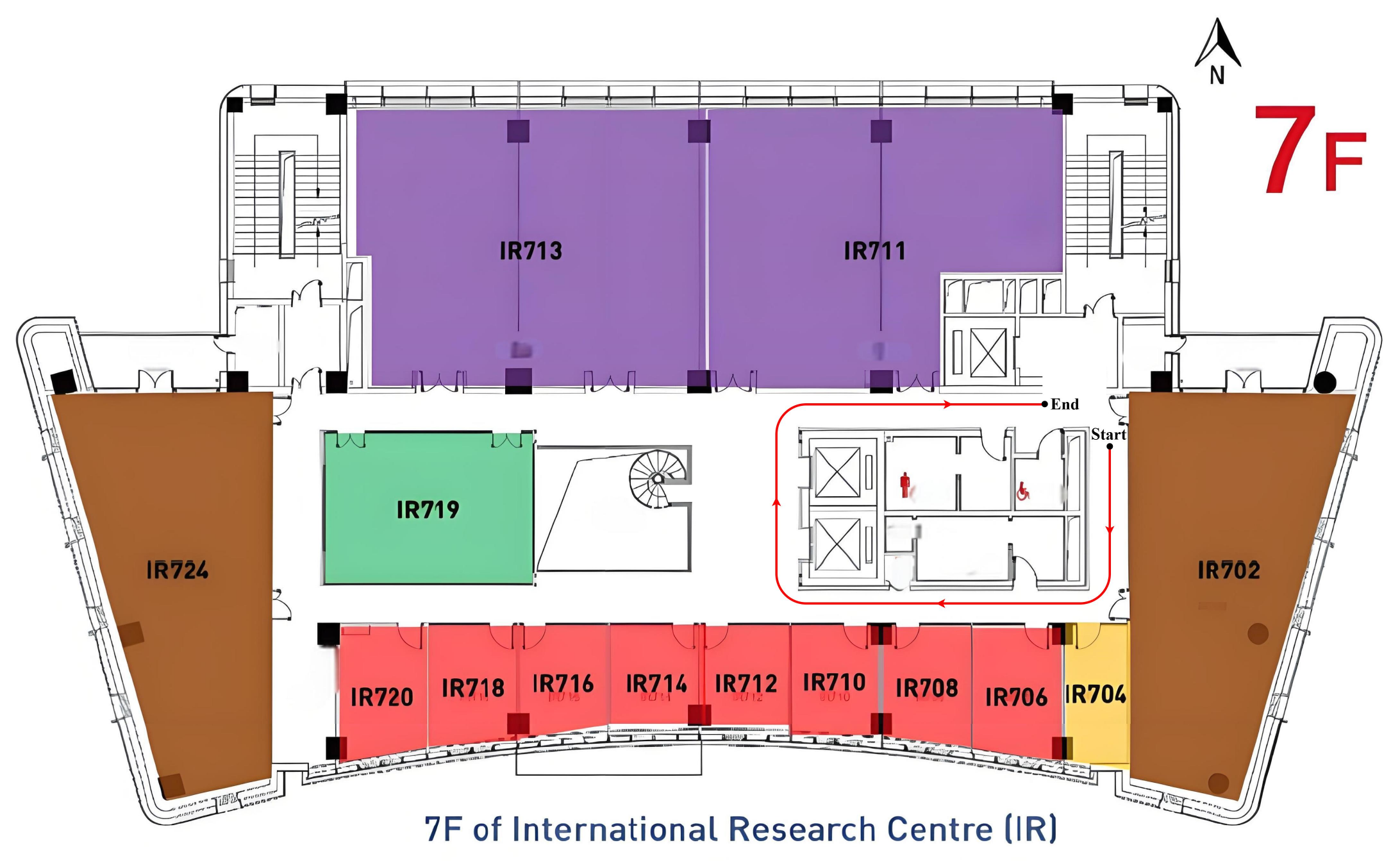}
  \caption{Experimental path with its start and end points on the 7th floor of
    the IR building.}
  \label{fig:exp-path}
\end{figure}

\begin{figure}[htbp]
  \centering%
  \subfloat[6th Floor.]{\includegraphics[width=0.8\linewidth]{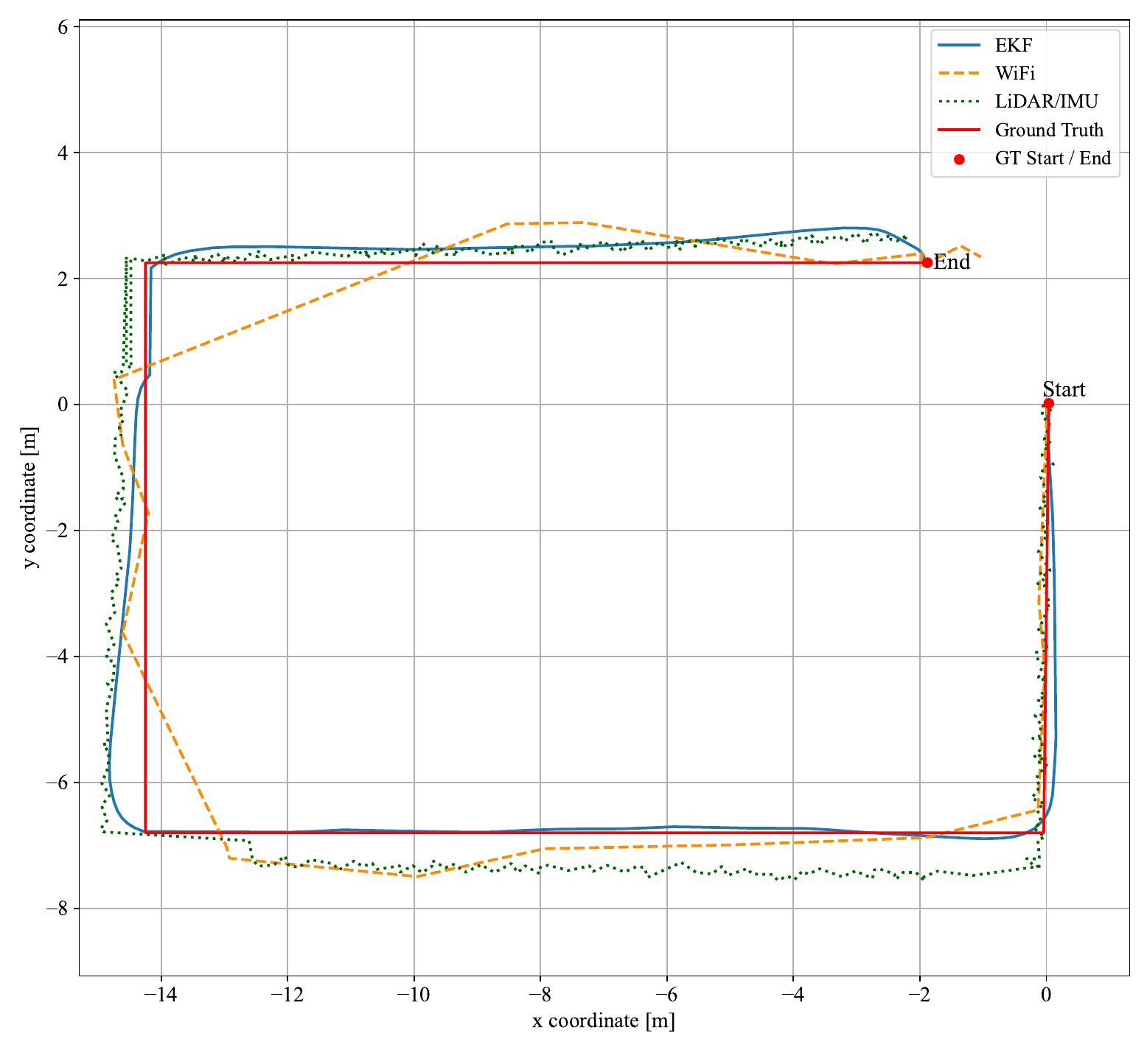}}\\
  \subfloat[7th Floor.]{\includegraphics[width=0.8\linewidth]{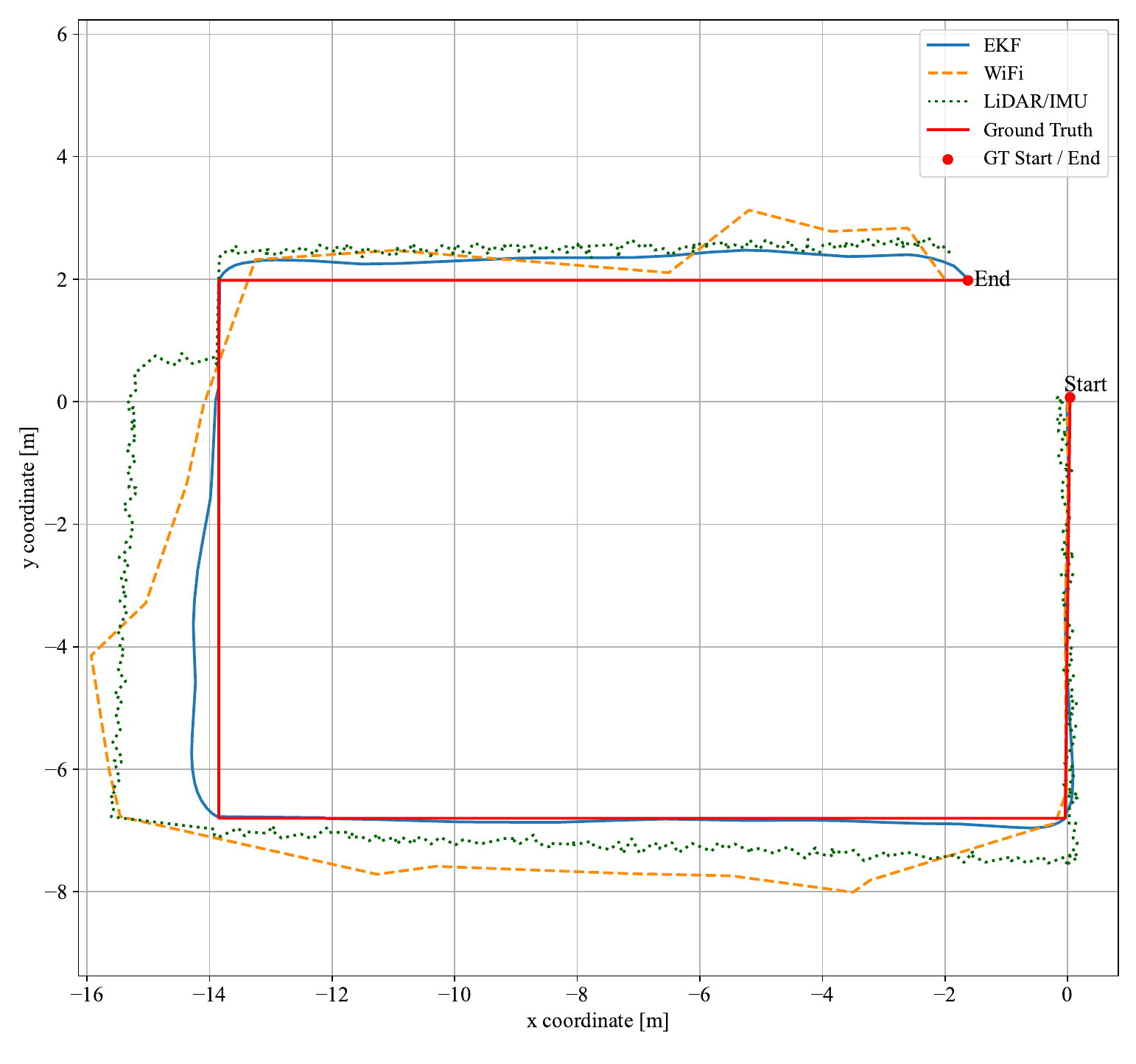}}\\
  \subfloat[8th Floor.]{\includegraphics[width=0.8\linewidth]{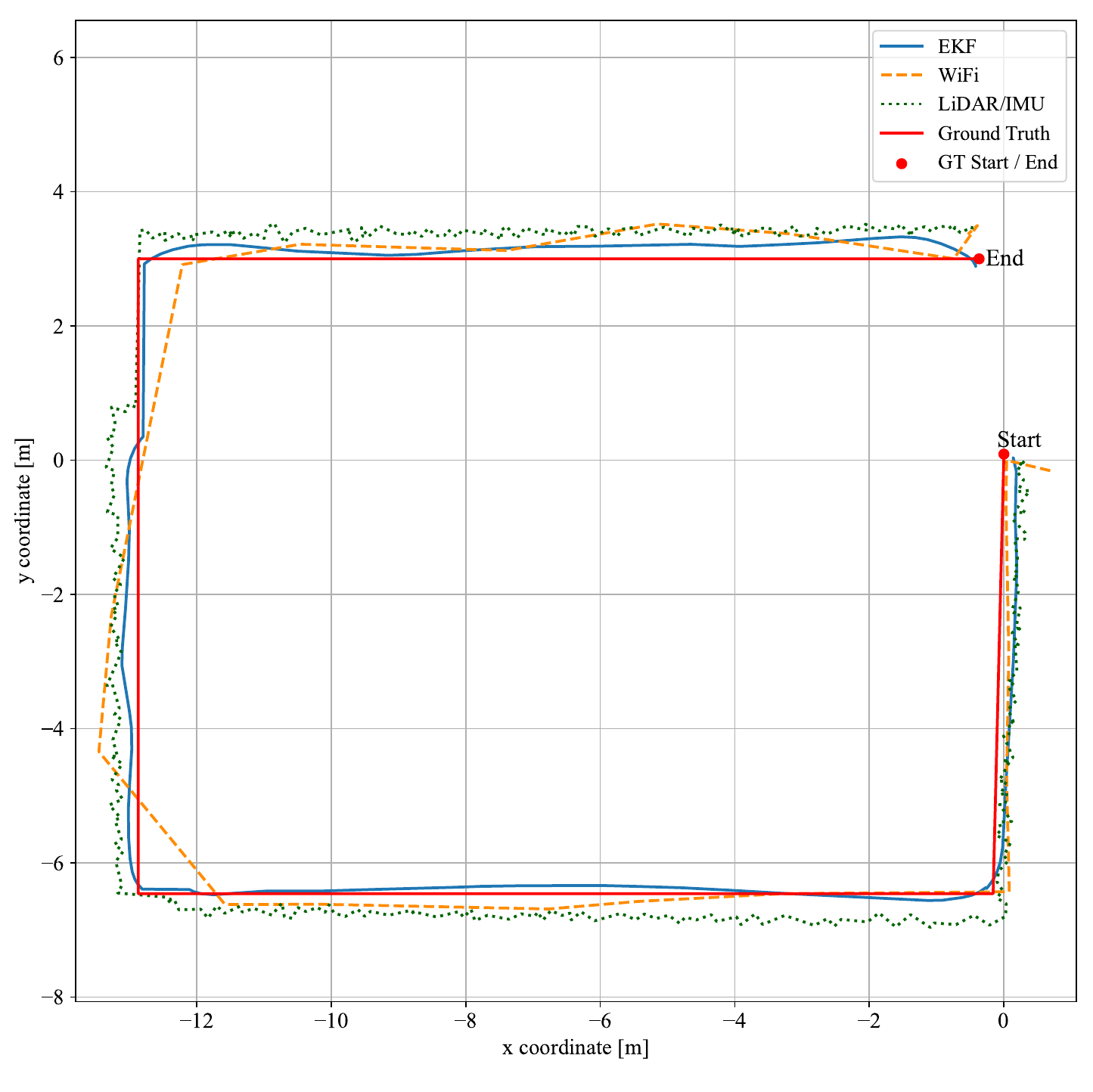}}
  \caption{Estimated trajectories on the floors of the IR building.}
  \label{fig:floor_comparison}
\end{figure}

The floor plans of the 6th, 7th, and 8th floors of the IR building are similar
to one another, whose example is shown in \autoref{fig:exp-path} for the
experimental path for the AGV on the 7th floor.
Repeating the same experiment on the three different floors, we obtained the
three maps shown in \autoref{fig:floor_comparison}~(a)--(c), each of which
includes the three estimated trajectories for Wi-Fi, LiDAR/IMU, and EKF together
with a true trajectory (i.e., ground truth).
The mean 2D errors of the nine trajectories (i.e., three methods on three
floors) are summarized in Table~\ref{tab:floor_comparison}. 

\autoref{fig:floor_comparison} shows that the estimated trajectories by the
proposed EKF-based multi-sensor fusion are more stable and closer to the true
trajectories in general than those by Wi-Fi and LiDAR/IMU, which is also
confirmed by the mean 2D errors summarized in
Table~\ref{tab:floor_comparison}. These experimental results demonstrate the
effectiveness of the EKF-based multi-sensor fusion in addressing the issues of
the drift and offset by Wi-Fi and LiDAR/IMU methods in indoor environments,
which prevents them from reliably estimating the exact positions of a moving
object.

Taking the 7th floor as an example, the mean 2D errors of Wi-Fi and LiDAR/IMU
are \SI{0.8792}{\m} and \SI{0.8553}{\m}, respectively, while that of EKF is only
\SI{0.2449}{\m}. The results on the 6th and 8th floors also show similar trends,
verifying that the multi-sensor fusion achieved by EKF can effectively reduce
localization errors and improve the system's robustness and stability against
different environmental disturbances.
\begin{table}[H]
  \centering
  \caption{Comparison of the mean 2D errors over multiple floors.}
  \label{tab:floor_comparison}
  \begin{tabular}{@{}ccc@{}}
    \toprule
    \textbf{Floor} & \textbf{Method} & \textbf{Mean 2D Error [\si{\m}]} \\
    \midrule
    \multirow{3}{*}{6th} 
                   & Wi-Fi & 0.5127 \\
                   & LiDAR/IMU & 0.6233 \\
                   & EKF & \textbf{0.3781} \\
    \midrule
    \multirow{3}{*}{7th} 
                   & Wi-Fi & 0.8792 \\
                   & LiDAR/IMU & 0.8553 \\
                   & EKF & \textbf{0.2449} \\
    \midrule
    \multirow{3}{*}{8th} 
                   & Wi-Fi & 0.8624 \\
                   & LiDAR/IMU & 0.6756 \\
                   & EKF & \textbf{0.2634} \\
    \bottomrule
  \end{tabular}
\end{table}

\subsection{Forward and Backward Trajectory Estimation}
\label{sec:forward-backward-exp}
\begin{figure}[H]
  \centering
  \subfloat[Forward~path.]{\includegraphics[width=0.9\linewidth]{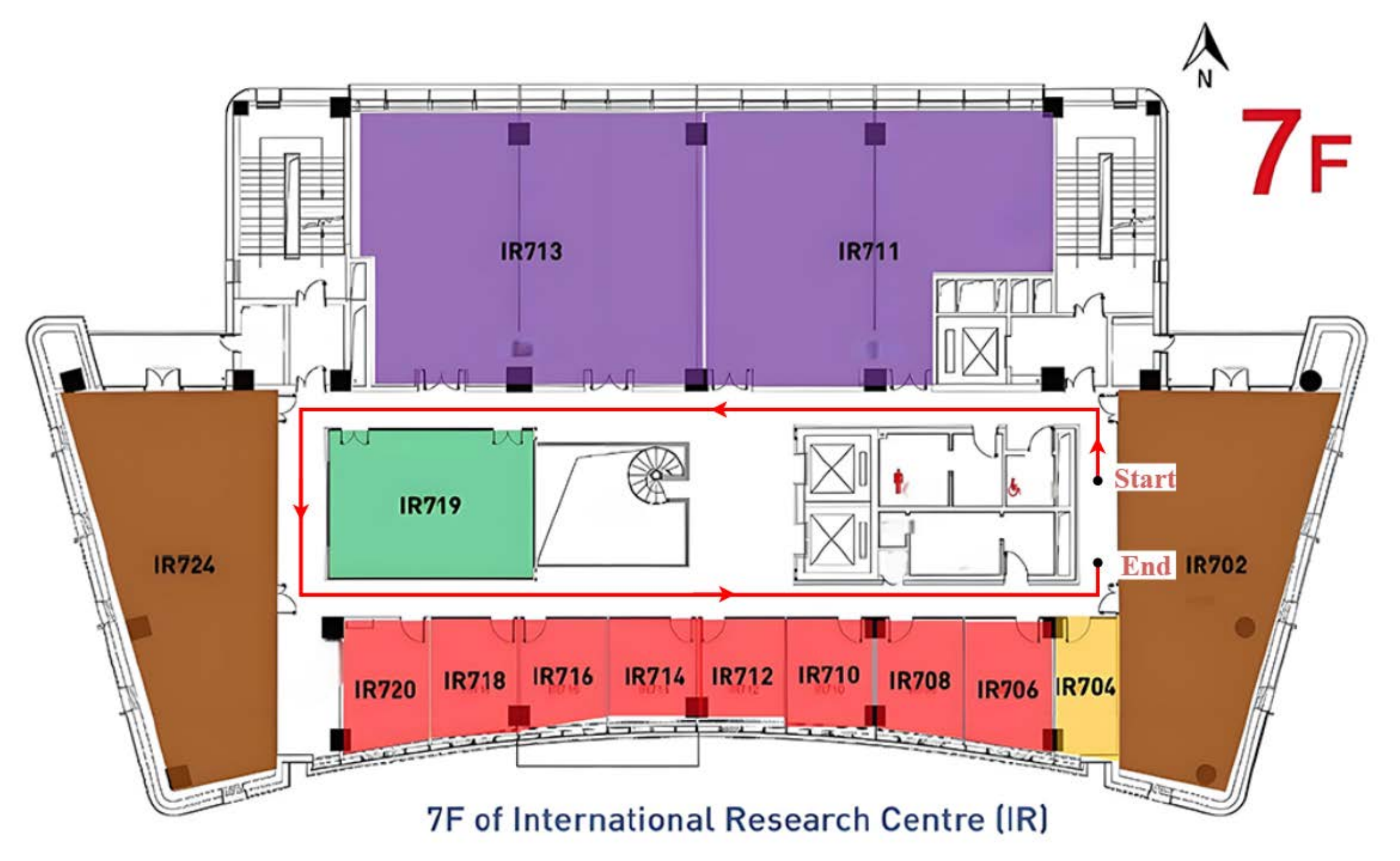}}\\
  \subfloat[Backward~path.]{\includegraphics[width=0.9\linewidth]{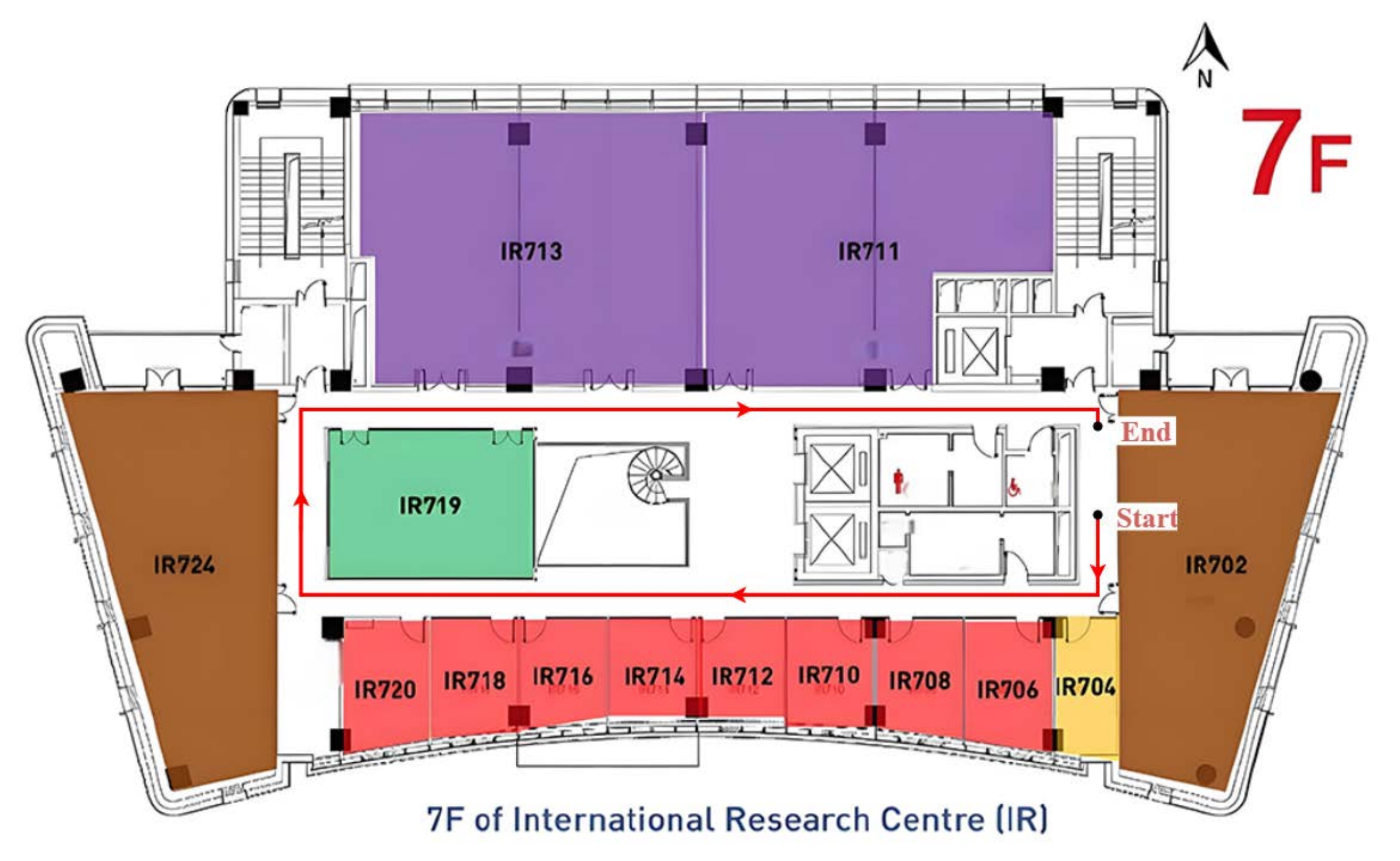}}
  \caption{Forward and backward paths with their start and end points on the 7th
    floor of the IR building.}
  \label{fig:forward-backward-paths}
\end{figure}

To further investigate the stability and adaptability of the implemented
localization system under different path settings, we conducted experiments for
forward and backward trajectory
estimation. \autoref{fig:forward-backward-paths}~(a) and (b) show the estimated
trajectories by the three localization methods for the forward (clockwise) and
backward (counterclockwise) paths on the 7th floor of the IR building,
respectively. The mean 2D errors of the six trajectories (i.e., three methods
for forward and backward paths) are summarized in
Table~\ref{tab:path_comparison}.

Like the results of the multi-floor trajectory estimation presented in
Section~\ref{sec:multi-floor-exp}, \autoref{fig:forward-backward-paths} shows
that the estimated trajectories, by the proposed EKF-based multi-sensor fusion,
are more stable and closer to the true trajectories in general than those by
Wi-Fi and LiDAR/IMU under the forward and backward path scenarios, which is
again confirmed by the mean 2D errors summarized in
Table~\ref{tab:path_comparison}. Though the system faces the differences in
motion patterns and the changes in observation distribution when moving along
the backward path, the EKF-based multi-sensor fusion can still maintain a
relatively lower estimation deviation, demonstrating its strong model
generalization capabilities.

Compared to EKF, Wi-Fi and LiDAR/IMU show significantly higher estimation
errors, especially right after the turns on the three corners, due to their
susceptibility to signal obstruction and inertial drift, which leads to
cumulative estimation bias; in the case of the proposed multi-sensor fusion
method, because EKF uses a prediction-correction mechanism to effectively fuse
global observations and local dynamics, it can successfully suppress cumulative
errors and thereby improve overall localization accuracy and robustness.

In addition, the comparison of the results for the forward and backward paths
reveals that the system exhibits good directional symmetry and path
invariance. These experimental results further demonstrate the capability of the
proposed EKF-based multi-sensor fusion method in handling different path
configurations and dynamic disturbances in real-world deployments.
\begin{table}[ht]
  \centering
  \caption{Comparison of the mean 2D errors over forward and backward paths.}
  \label{tab:path_comparison}
  \begin{tabular}{@{}ccc@{}}
    \toprule
    \textbf{Path} & \textbf{Method} & \textbf{Mean 2D Error [\si{\m}]} \\
    \midrule
    \multirow{3}{*}{Forward} 
                  & Wi-Fi & 1.3404 \\
                  & LiDAR/IMU & 2.8803 \\
                  & EKF & \textbf{0.3672} \\
    \midrule
    \multirow{3}{*}{Backward} 
                  & Wi-Fi & 1.2651 \\
                  & LiDAR/IMU & 1.8968 \\
                  & EKF & \textbf{0.2851} \\
    \bottomrule
  \end{tabular}
\end{table}
\begin{figure}[ht]
  \centering
  \subfloat[Forward~path.]{\includegraphics[width=0.9\linewidth]{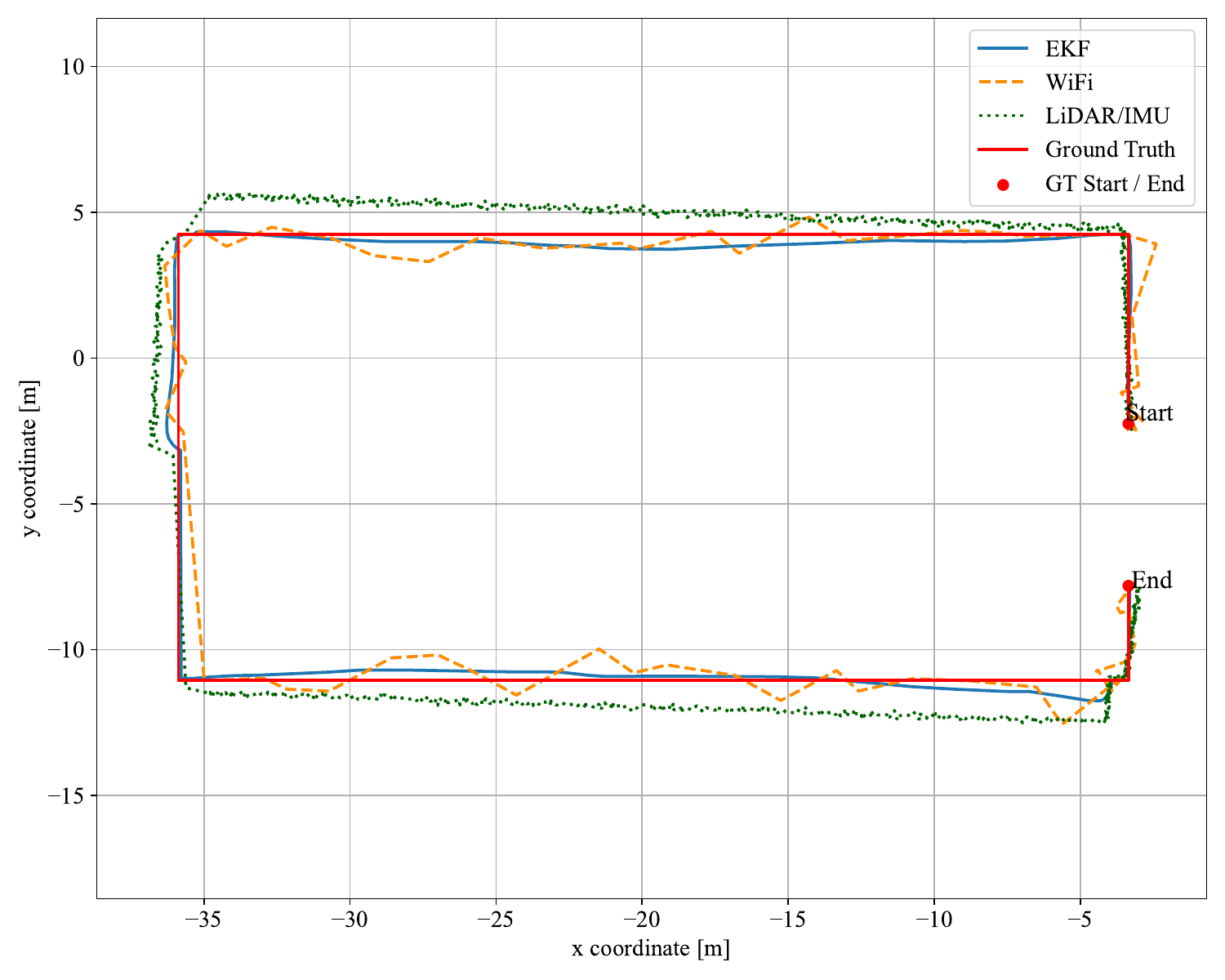}}\\
  \subfloat[Backward~path.]{\includegraphics[width=0.9\linewidth]{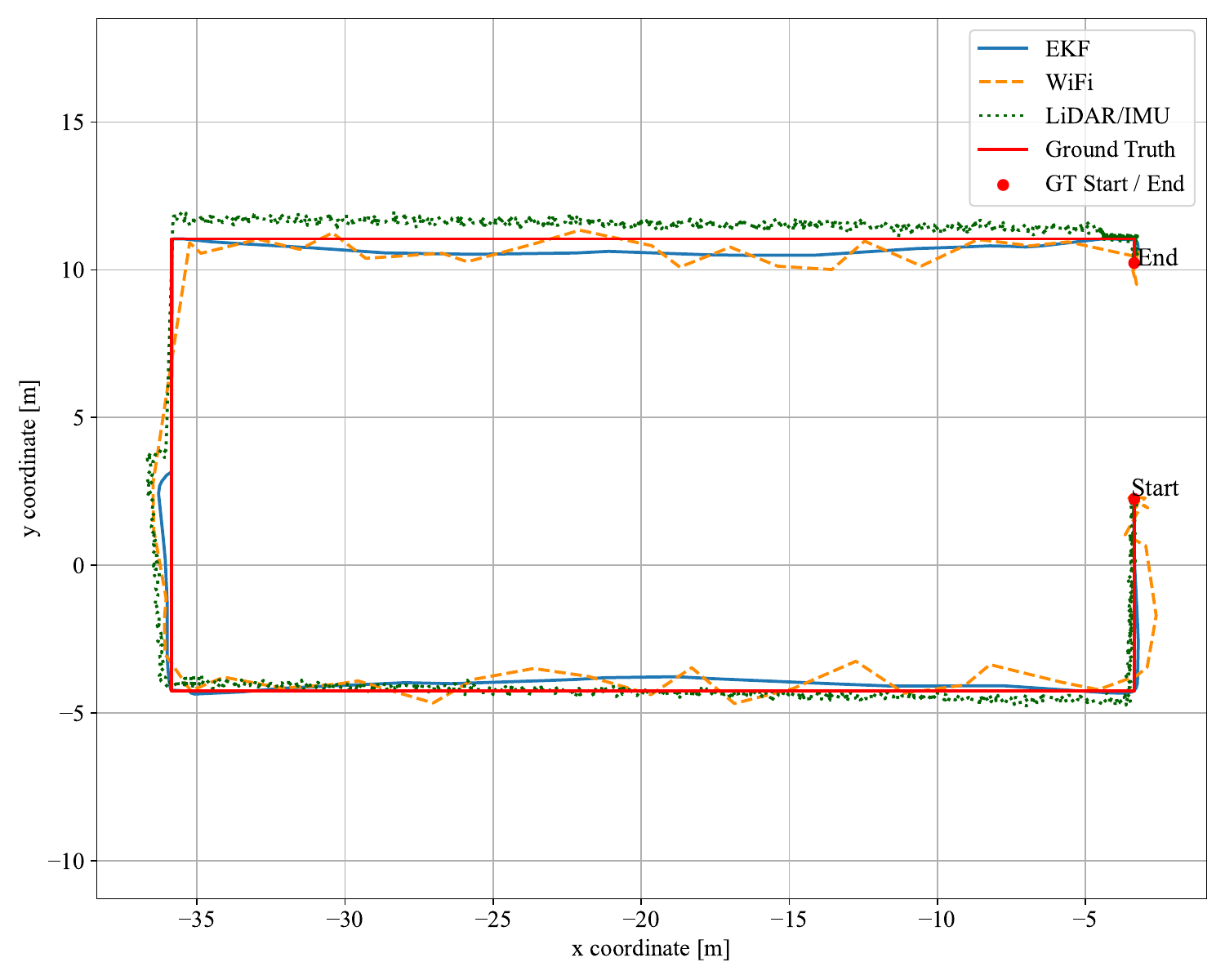}}
  \caption{Estimated trajectories for the forward and backward paths on the 7th
    floor of the IR building.}
  \label{fig:path_comparison}
\end{figure}

\section{Conclusions}
\label{sec:conclusions}
We have proposed an indoor localization and navigation framework that integrates
DNN-based Wi-Fi RSSI fingerprinting, IMU-assisted dynamic localization, and
LiDAR environmental perception capabilities through multi-sensor fusion based on
the EKF algorithm. We implemented the proposed localization system on an AGV and
conducted comprehensive field experiments with it on the 6th, 7th, and 8th
floors of the IR building at XJTLU to verify the stability and localization
accuracy of the proposed EKF-based multi-sensor fusion method under multipath
interference and signal drift.

The experimental results demonstrate that the multi-sensor fusion method based
on the EKF maintains high positioning accuracy under various path conditions and
floor environments, with the mean 2D error consistently controlled between
\SI{0.2449}{\m} and \SI{0.3781}{\m}. In contrast, the DNN-based Wi-Fi regression
model exhibits significant error fluctuations, with errors ranging from
\SI{0.5127}{\m} to \SI{1.3404}{\m} under signal obstruction or multipath
effects. The LiDAR/IMU-based dynamic positioning, too, exhibits greater
positioning errors distributed between \SI{0.6233}{\m} and \SI{2.8803}{\m} due
to the IMU cumulative error issues. By fusing multi-source sensor data, the EKF
effectively suppresses the deviations and uncertainties caused by the
aforementioned error sources, making the overall trajectory more closely aligned
with the actual path. This significantly enhances the system's robustness and
adaptability in complex environments.

\section*{Acknowledgment}
This work was supported in part by Xi'an Jiaotong-Liverpool University (XJTLU)
Summer Undergraduate Research Fellowships (under Grant SURF-2024-0311).

\balance 


\end{document}